\documentclass[pdflatex,sn-mathphys-num]{sn-jnl} 

\usepackage[utf8]{inputenc}
\usepackage[T1]{fontenc}
\usepackage{amsmath,amsfonts}
\usepackage{algorithm}
\usepackage{algpseudocode}
\usepackage{array}
\usepackage{graphicx}
\usepackage{subcaption}
\usepackage{url}
\usepackage{verbatim}
\usepackage[export]{adjustbox}
\usepackage{booktabs}

\def\F{{\mathbf{F}}} 

\def\J{{\mathcal{J}}}
\def\G{{\mathcal{G}}}

\newcommand{\Fmetric}{\mathcal{F}}   

\begin{document}

\title{Attention-Guided Cross-Temporal Clustering for Self-Supervised Video Object Segmentation}

\author[1]{\fnm{Waqas} \sur{Arshid}}\email{waqas.arshid@griffith.edu.au}

\author*[1]{\fnm{Mohammad} \sur{Awrangjeb}}\email{mohammad.awrangjeb@griffith.edu.au}

\author[1]{\fnm{Alan Wee-Chung} \sur{Liew}}\email{a.liew@griffith.edu.au}

\author[2]{\fnm{Yongsheng} \sur{Gao}}\email{yongsheng.gao@griffith.edu.au}

\affil[1]{\orgdiv{School of Information and Communication Technology}, \orgname{Griffith University}, \city{Brisbane}, \state{QLD}, \country{Australia}}
\affil[2]{\orgdiv{School of Engineering and Built Environment -- Electrical and Electronic Engineering}, \orgname{Griffith University}, \city{Brisbane}, \state{QLD}, \country{Australia}}

\abstract{
Video object segmentation (VOS) is a fundamental task in video understanding, requiring accurate delineation and consistent tracking of objects across frames. While supervised methods achieve strong performance, they depend on densely annotated datasets that are costly to obtain and limited in domain coverage. Self-supervised learning offers a promising alternative by removing the need for manual labels; however, existing approaches often struggle to jointly maintain spatial accuracy and temporal coherence, particularly in unconstrained multi-object scenarios. Many rely on optical flow, synthetic motion cues, or task-specific pretraining, limiting scalability and generalisation. We propose a self-supervised framework, Cross-Temporal Consistency and Clustering (CTC\textsuperscript{2}), that learns mid-level, part-aware representations by combining attention-guided token selection with lightweight temporal clustering.
Instead of operating at the pixel or whole-object level, the method aligns soft part assignments across time using a saliency-weighted symmetric consistency objective. The framework leverages a frozen transformer backbone with lightweight modules for adaptive token selection and multi-offset temporal alignment, enabling efficient scaling across resolutions and motion patterns.
CTC\textsuperscript{2} achieves competitive performance among recent self-supervised methods while maintaining real-time throughput, without relying on motion cues or domain-specific adaptation. It further demonstrates stable behaviour under cross-dataset evaluation and can be extended to a semi-supervised setting using a first-frame mask. These results suggest that attention-guided token selection combined with temporal clustering offers a practical and scalable direction for label-free video segmentation.
}


\maketitle

\begin{center}
\small
Accepted for publication in \textit{Machine Intelligence Research}. 
DOI: 10.1007/s11633-026-1648-7.
\end{center}

\section{Introduction}

Understanding how objects evolve over time—how they move, deform, interact, or become occluded—is central to visual intelligence. Video Object Segmentation (VOS) supports applications such as autonomous navigation, intelligent video editing, augmented reality, and human--robot interaction~\cite{miao2024discriminative, perazzi2016benchmark}. While supervised methods achieve strong performance using dense per-frame annotations~\cite{xu2018youtube, maninis2018video}, they scale poorly: pixel-accurate labels remain costly to obtain at scale, and large annotated datasets are not always available for new domains or rare categories. This bottleneck limits deployment in settings with privacy constraints, long-tail categories, or rapidly changing domains.

To mitigate annotation dependence, recent work has increasingly explored \emph{self-supervised} VOS, which aims to learn spatiotemporal representations directly from unlabeled videos~\cite{lai2019self, vondrick2018tracking, salehi2023time}. Many methods replace human labels with intrinsic signals from temporal coherence, motion consistency, or appearance regularities. However, existing approaches vary widely in the assumptions they require, ranging from motion-based cues and synthetic labels to correspondence learning frameworks that rely on strong data augmentations or contrastive objectives. Common dependencies include optical flow~\cite{wang2019learning}, which can fail in low-texture or fast-motion regions; motion saliency~\cite{yang2021collaborative}, which is often confounded by background dynamics; and pseudo-labels from synthetic cues~\cite{huang2025segment, zhu2023tracking}, which may introduce bias.
Meanwhile, correspondence-based SSL methods—such as CRW, TimeT, and TAPIR-style matching—show that flow-free alignment is feasible, but may still break under occlusion, clutter, or large appearance changes.
These limitations motivate dense video representations that remain temporally consistent under viewpoint change, occlusion, and background noise—without brittle external priors.

We argue that a promising direction is to learn representations at the level of \emph{semantic object parts}, rather than whole objects or individual pixels. Parts such as wheels, wings, or ears tend to persist through occlusion, deform predictably, and compose into objects~\cite{ziegler2022self, wei2024towards, zheng2024surface}. This mid-level abstraction generalises across categories and can be more stable under viewpoint change and intra-class variation~\cite{hung2019scops, xu2022groupvit}. However, discovering temporally consistent parts without supervision remains challenging, particularly when neither motion cues nor semantic annotations are available. Prior work has explored unsupervised part discovery in static images and short-range correspondences in videos, but extending these ideas to long-range, label-free temporal consistency remains an open problem.

In this work, we build upon clustering-based part discovery and temporal correspondence learning to develop a self-supervised framework that encourages stable part-level alignment across time. We match salient spatial tokens across adjacent frames and enforce consistency in their soft cluster assignments. The key intuition is that the same semantic part (e.g., a bear’s ear or a car’s wheel) should induce similar feature distributions over time, even under scale, rotation, or partial occlusion~\cite{kim2010rotation}. We operationalise this intuition using a symmetric Kullback--Leibler (KL) consistency loss, chosen for balanced alignment rather than as a new objective, and we provide ablations against one-sided KL, cross-entropy, and contrastive alternatives.

A central component is the \texttt{[CLS]} attention map from a frozen SAM2 encoder~\cite{ravi2024sam} to identify salient tokens. Rather than introducing a new saliency mechanism, we leverage an emergent property of ViTs: the \texttt{[CLS]} token aggregates global context and often highlights semantically meaningful regions. In practice, strong \texttt{[CLS]}-attention responses align with salient regions~\cite{caron2021emerging, zheng2024surface}. We adopt an adaptive top-$p$ selection strategy with coarse grid diversity to balance semantic focus and spatial coverage, mitigating failure cases involving small objects, clutter, and varying resolutions.

Technically, we treat the frozen SAM2 encoder as a general-purpose visual tokenizer trained on large-scale segmentation data~\cite{kirillov2023segment}. Token embeddings are fed to a lightweight MLP clustering head to produce soft part assignments. We track clusters using cosine-similarity matching and enforce temporal alignment via the symmetric KL objective. Keeping the backbone frozen reduces overhead and avoids domain-specific fine-tuning; however, we explicitly evaluate frozen versus fine-tuned variants. To handle variable motion and frame rates, we introduce a \emph{multi-$\Delta t$ temporal pyramid} with match-rate control, inspired by multi-step correspondence strategies but tailored to stable clustering rather than contrastive tracking.

Our approach complements prior work such as SelfMask~\cite{yang2021collaborative}, TimeT~\cite{salehi2023time}, BETRayed Attention~\cite{ding2024betrayed}, and modern flow-free correspondence models. Rather than relying on motion cues, heavy augmentations, or decoder heads, CTC\textsuperscript{2} focuses on attention-guided token selection and cross-temporal part clustering. This mid-level grouping improves robustness to occlusion, background interference, and intra-class variability. We also relate our design to memory-centric VOS methods such as XMem~\cite{cheng2022xmem} and AOT~\cite{yang2021associating}, while emphasising that our framework does not use external memory modules or mask propagation.

We evaluate on three established self-supervised VOS benchmarks—DAVIS-2017~\cite{ponttuset2017davis}, DAVIS-2016~\cite{perazzi2016benchmark}, and YouTube-VOS~\cite{xu2018youtube}. Following common practice in label-free dense segmentation~\cite{hung2019scops, ziegler2022self}, the first annotated frame is used \emph{only} to define the evaluation mapping. For completeness, we also report semi-supervised results where the first-frame mask is provided at inference, and cross-dataset generalisation experiments to assess robustness under domain shift. Across these settings, CTC\textsuperscript{2} achieves competitive performance while maintaining real-time throughput.

In summary, our main contributions are:
\begin{itemize}
 \item We present a self-supervised framework for \emph{part-level temporal consistency} that combines attention-guided token selection with lightweight cross-temporal clustering and a symmetric KL consistency objective.
 \item We introduce adaptive token budgeting based on top-$p$ attention and spatial diversity, improving semantic coverage across resolutions without increasing computational complexity.
 \item We develop a multi-$\Delta t$ temporal supervision strategy with match-rate control to improve correspondence stability under variable motion and frame rates.
 \item We report competitive results on DAVIS-2017, DAVIS-2016, and YouTube-VOS, supported by cross-dataset and semi-supervised evaluations highlighting robustness.
\end{itemize}

\section{Related Work}
\label{sec:related_work}

\subsection{Self-Supervised Learning for Dense Visual Understanding.}
Self-supervised learning (SSL) has become a foundational paradigm in computer vision, enabling representation learning without human-annotated labels~\cite{jing2020self}. Early image-level methods such as MoCo~\cite{he2020momentum}, BYOL~\cite{grill2020bootstrap}, and SimSiam~\cite{chen2020exploring} established instance discrimination as a strong pretext task, achieving competitive performance in classification and retrieval. However, because these approaches emphasise global embeddings, transferring them directly to dense prediction tasks—where spatial correspondence is essential—remains challenging.

To address this limitation, later work introduced spatial structure into SSL objectives. DenseCL~\cite{wang2021dense} aligns pixel-level features across augmented views; PixPro~\cite{xie2021propagate} propagates local features across spatial neighbourhoods. Complementary to pixel-centric objectives, \emph{part-aware} models such as SCOPS~\cite{hung2019scops} and Leopart~\cite{ziegler2022self} promote the discovery of spatially coherent parts, while GroupViT~\cite{xu2022groupvit} and TokenCut~\cite{wang2023tokencut} show that semantic grouping can \emph{emerge} from transformer tokenisation. DINO~\cite{caron2021emerging} further demonstrates that Vision Transformer attention provides an unsupervised saliency prior correlated with object and part structure, motivating attention-guided grouping.

More recent unsupervised grouping and segmentation methods extend these ideas. LOST~\cite{simeoni2021localizing} shows that object localization can emerge directly from self-supervised ViT features through attention-guided region extraction, while FreeSOLO~\cite{wang2022freesolo} and U2Seg~\cite{niu2024unsupervised} generalise grouping to class-agnostic instance and universal image segmentation. These methods highlight the strength of transformer representations for dense grouping without supervision, but operate primarily on static images and do not enforce temporal coherence across frames. Dense unsupervised video segmentation~\cite{araslanov2021dense} begins to address temporal consistency, yet remains pixel-centric and does not leverage mid-level part representations or attention-guided token selection.

We build on these insights in two ways: (i) instead of clustering \emph{all} tokens or pixels, we exploit ViT attention to \emph{selectively} retain salient spatial tokens from a frozen encoder using an \emph{adaptive} budget that preserves coverage across resolutions; and (ii) we extend grouping from static images to \emph{videos} by enforcing \emph{cross-temporal} consistency of soft part assignments through a lightweight alignment mechanism.

\subsection{Temporal Self-Supervision in Video Representation Learning.}
Temporal continuity in video provides a rich source of self-supervision~\cite{das2023towards}. Early approaches use surrogate objectives such as frame ordering (Shuffle and Learn~\cite{misra2016shuffle}, Arrow of Time~\cite{wei2018learning}) and future prediction~\cite{vondrick2016anticipating}. Cycle-consistency later emerged as a powerful constraint: TimeCycle~\cite{dwibedi2019temporal} enforces that correspondences traced forward in time return to their origin, while subsequent methods operationalise correspondence via contrastive or tracking-based objectives, including CycleContrast~\cite{wang2022cyclecontrast}, spacetime random walks~\cite{jabri2020space} and memory-augmented tracking (MAST)~\cite{lai2020mast}. Time-tuning~\cite{salehi2023time} further adapts image-pretrained features to unlabeled video using temporal alignment losses.

Despite their effectiveness, many of these methods either (i) operate on \emph{global} embeddings that obscure fine-grained structure, or (ii) rely on dense warping or optical flow, which can be fragile under fast motion, occlusion, or low texture and computationally expensive at scale.
Recent correspondence-based models such as TAPIR~\cite{doersch2023tapir} demonstrate strong long-range, flow-free tracking through recurrent refinement, while Deep ViT Feature Descriptors~\cite{amir2021deep} show that dense correspondences can emerge from self-supervised ViTs without explicit motion modeling. However, these methods still operate on densely sampled patches and do not explicitly constrain the temporal stability of part-level structures.

In contrast, we pursue \emph{token-level} temporal supervision. Guided by attention-derived saliency, we restrict learning to a compact, \emph{adaptively} selected set of informative tokens and align their \emph{soft} part assignments across time. This design avoids dense warping, relies on lightweight cosine-similarity matching, and supports a multi-$\Delta t$ temporal pyramid with match-rate control.
Rather than competing with dense correspondence methods such as TAPIR or ViT-based tracking, our formulation targets a lightweight, mid-level alternative in which part-like token groups—rather than pixels or dense patches—serve as the unit of temporal alignment.
We treat attention-derived saliency (e.g., \texttt{[CLS]} attention) as a heuristic prior rather than a definitive foreground indicator, acknowledging its limitations for small objects and cluttered scenes.

\subsection{Part-Level Representation Learning.}
Part-level representations provide semantically meaningful, generalisable units that are robust to occlusion, deformation, and viewpoint change~\cite{wei2024towards, alayrac2022flamingo}. In images, SCOPS~\cite{hung2019scops} and Leopart~\cite{ziegler2022self} discover recurring part structure through spatial grouping. In videos, temporal alignment of latent structures—such as cycle-consistency in TimeCycle~\cite{dwibedi2019temporal} and neural-surface-based PartDistillation~\cite{zheng2024surface}—suggests that mid-level components can persist across time.
Dense tracking works such as TAPIR~\cite{doersch2023tapir} and Deep ViT Correspondences~\cite{amir2021deep} further indicate that temporally stable mid-level features can emerge without flow, though they do not explicitly cluster tokens into part-like groups.

We extend this line of work by \emph{integrating transformer-derived token saliency} with a \emph{temporal clustering loss}. Salient tokens, selected via \texttt{[CLS]} attention from a frozen encoder, are softly clustered into parts and aligned across frames using a symmetric KL objective.
This encourages temporally stable groupings without assuming that clusters always correspond to fully interpretable semantic parts.

\subsection{Foundation Models and Saliency-Guided Learning.}
Large pretrained Vision Transformers, including DINO~\cite{caron2021emerging}, SAM~\cite{kirillov2023segment}, and SAM2~\cite{ravi2024sam}, exhibit strong transfer to dense prediction tasks due to hierarchical token representations and global attention. These models often show emergent grouping of semantically related regions, providing a natural cue for unsupervised segmentation. Leveraging \texttt{[CLS]} attention~\cite{dosovitskiy2020image} as a saliency signal enables extraction of informative tokens without training a separate saliency network. Our method adopts this principle using a frozen SAM2 encoder.
We emphasise that freezing the encoder is a design choice motivated by efficiency and generalisation rather than conceptual novelty; ablations in Sec.~4.3 examine frozen versus fine-tuned variants.

\subsection{Token-Level Consistency and Loss Design.}
Temporal stability in learned representations has been pursued through contrastive objectives (InfoNCE~\cite{oord2018representation}), cycle-consistency~\cite{wang2022cyclecontrast}, and patch-tracking losses~\cite{jabri2020space}, often requiring careful negative sampling or memory banks. We instead adopt a symmetric KL divergence~\cite{ziegler2022self} between matched token clusters, avoiding negative sampling while enforcing bidirectional consistency.
Prior work suggests that symmetric divergences can mitigate directional bias, though their advantages over one-sided KL, cross-entropy, or contrastive losses depend on the alignment setting; we therefore include ablations to clarify their behaviour in our framework.

Prior SSL video methods emphasise pixel- or object-level propagation, optical flow, or external memory. In contrast, our method treats \emph{parts} as the unit of temporal alignment, unifying attention-guided token selection with cross-temporal clustering and symmetric-KL agreement on a frozen backbone. With \emph{adaptive} token budgeting and a \emph{multi-$\Delta t$} training scheme, the pipeline remains label-free and decoder-free while scaling across resolutions and motion regimes.

\section{Methodology}
\label{sec:method}

\begin{table}[t] \centering \caption{Summary of notation used in Sec.~\ref{sec:method}.} \label{tab:symbols} \begin{tabular}{ll} \toprule Symbol & Meaning \\ \midrule $H,W$ & Spatial height and width of feature map \\ $N$ & Number of spatial tokens ($N=HW$) \\ $D$ & Token embedding dimension \\ $\mathbf{X}_t \in \mathbb{R}^{N\times D}$ & Token embeddings for frame $I_t$ \\ $\boldsymbol{\alpha}_t \in \mathbb{R}^N$ & Saliency prior from \texttt{[CLS]} attention \\ $\mathcal{S}_t$ & Indices of selected tokens after adaptive sampling \\ $k_t = |\mathcal{S}_t|$ & Number of selected tokens \\ $\mathbf{P}_t(i) \in \Delta^K$ & Soft part distribution for token $i$ \\ $K$ & Number of part prototypes \\ $\mathcal{M}_{t,\Delta}$ & Mutual matches between frames $t$ and $t+\Delta$ \\ \bottomrule \end{tabular} \end{table}

\subsection{Overview and Notation}

Our objective is to learn temporally consistent and semantically coherent \emph{part-level} representations of objects in videos in a purely self-supervised manner. Reasoning at the level of constituent parts—such as wheels, limbs, or articulated components—has been shown to provide stronger inductive biases than whole-object or pixel-level formulations, particularly under occlusion and deformation~\cite{hung2019scops, aubret2024self}. This perspective complements our earlier CAMVOS framework~\cite{arshid2024camvos}, which emphasised contextual and long-term memory for video object segmentation. In contrast, the present work adopts an explicitly \emph{unsupervised, part-centric} formulation that discovers latent part groupings and enforces their temporal consistency across a video sequence.

At a high level, the framework consists of four stages. First, a frozen encoder produces dense spatial tokens together with an attention-derived saliency prior. Second, an adaptive token selection strategy extracts a compact yet semantically informative subset of tokens. Third, a lightweight MLP-based clustering head maps each selected token to a soft distribution over $K$ latent parts. Finally, a temporal consistency objective aligns these part distributions across multiple time offsets.

For an input frame $I_t$, the encoder outputs a feature map $\mathbf{F}_t \in \mathbb{R}^{H \times W \times D}$, which is flattened into $N = H \times W$ spatial tokens $\mathbf{X}_t \in \mathbb{R}^{N \times D}$. A saliency score $\alpha_t^i$ is assigned to each token via the encoder’s \texttt{[CLS]} attention, forming the vector $\boldsymbol{\alpha}_t \in \mathbb{R}^N$. All notation is summarised in Table~\ref{tab:symbols}.

An adaptive sampling rule (Sec.~\ref{subsec:adaptive_selection}) selects a subset $\mathcal{S}_t \subseteq \{1,\dots,N\}$ of size $k_t$. Each selected token $i \in \mathcal{S}_t$ is mapped to a soft part distribution $\mathbf{P}_t(i) \in \Delta^K$, where $K$ denotes the number of latent part prototypes. Temporal correspondence is established via cosine similarity between tokens from frames $t$ and $t+\Delta$, yielding the mutual match set $\mathcal{M}_{t,\Delta} \subseteq \mathcal{S}_t \times \mathcal{S}_{t+\Delta}$.

To ground the notation, consider a concrete example. A $224 \times 224$ frame yields $N=196$ tokens using a patch size of 16. The saliency prior selects $k_t=32$ informative tokens. After clustering, each token is assigned a $K=8$-dimensional part distribution. Matching these tokens with those from frame $t{+}3$ produces 24 mutual correspondences, on which the symmetric distributional alignment loss is computed.

Our design is guided by three principles. First, \emph{efficiency through adaptive saliency}: selecting only salient and spatially diverse tokens reduces computation while preserving semantic coverage. Second, \emph{stability through part-level clustering}: grouping tokens into soft parts yields representations that are more robust to occlusion, motion, and viewpoint change than pixel-level alignment. Third, \emph{temporal consistency via symmetric alignment}: we employ symmetric KL divergence as a simple, bidirectional distribution-matching objective. Together, these principles define a lightweight and fully unsupervised pipeline for video object segmentation.

\begin{figure*}[!t]
 \centering
 \includegraphics[width=\textwidth, height=0.5\textheight,keepaspectratio]{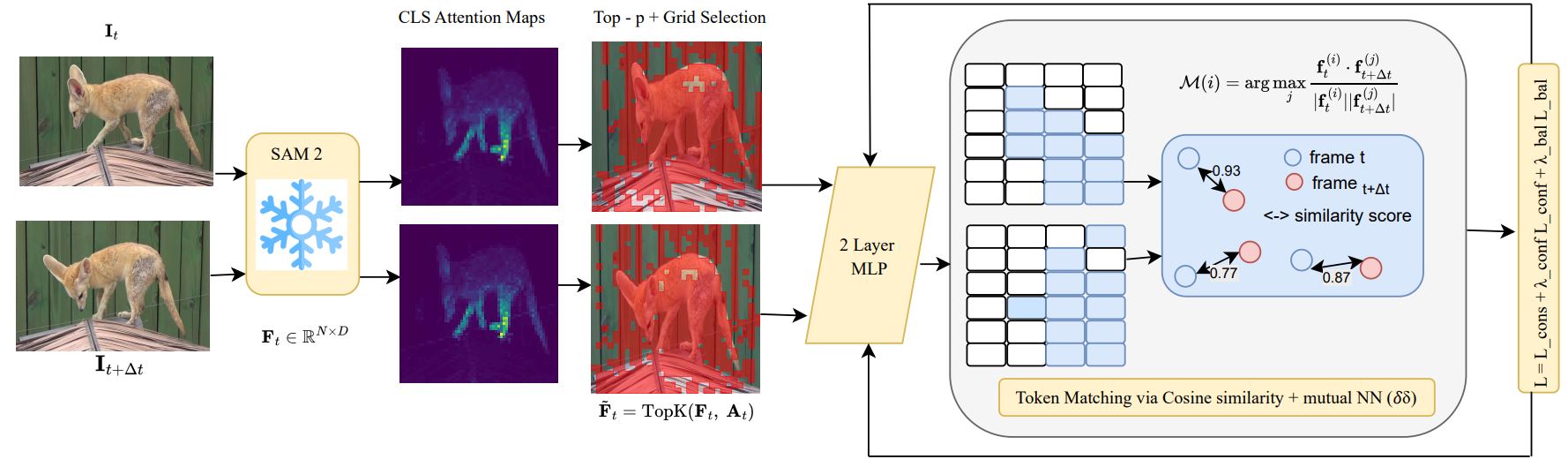}
 \caption{Overview of the proposed CTC\textsuperscript{2} framework.
Consecutive frames $\mathbf{I}t$ and $\mathbf{I}{t+\Delta t}$ are encoded by a frozen SAM2 ViT backbone. The \texttt{[CLS]} attention map guides adaptive token selection, retaining only the most salient and spatially diverse regions. A lightweight MLP head clusters these tokens into soft part assignments. Cross-temporal matching aligns part distributions via cosine similarity across multiple temporal offsets, while a saliency-weighted symmetric KL divergence enforces temporal consistency.}
 \label{fig:methodology}
\end{figure*}

\subsection{Encoder and Saliency Prior}

We adopt a frozen SAM2 encoder as the feature extractor, motivated by its strong ability to capture high-level semantic information across diverse visual domains. For each frame $I_t$, the encoder produces a dense feature map $\mathbf{F}_t \in \mathbb{R}^{H \times W \times D}$, which is flattened into $N=H \times W$ spatial tokens $\mathbf{X}_t = \{\mathbf{x}_t^i\}_{i=1}^{N}$ with $\mathbf{x}_t^i \in \mathbb{R}^D$.

In addition to spatial embeddings, the encoder exposes a global \texttt{[CLS]} token whose final-layer attention provides an importance weight for each spatial location. We treat this signal as a saliency prior $\boldsymbol{\alpha}_t \in \mathbb{R}^N$.

The \texttt{[CLS]} attention offers two practical advantages. First, it provides a training-free mechanism for highlighting semantically meaningful regions without introducing an additional saliency network or optical-flow estimator. Second, by keeping the encoder frozen, optimisation is confined to the lightweight clustering head and temporal alignment modules, reducing training complexity.

We note that this saliency prior is an imperfect heuristic: \texttt{[CLS]} attention may be unreliable in cluttered scenes, for small objects, or under domain shift. To mitigate this, our adaptive selection module combines saliency weighting with spatial diversity constraints, ensuring broad coverage even when the prior is noisy.

\subsection{Frame Encoding and Attention-Derived Saliency}
\label{subsec:encoding_saliency}

Given a video frame $\mathbf{I}_t \in \mathbb{R}^{3 \times H \times W}$, we extract patch-level embeddings using a frozen SAM2 ViT encoder~\cite{jiaxing2025sam2}. The frame is divided into $N=\tfrac{H}{P}\cdot\tfrac{W}{P}$ patches of size $P{\times}P$ (with $P{=}16$), producing a feature map $\mathbf{F}_t \in \mathbb{R}^{N \times D}$ with embedding dimension $D{=}768$.
Tokens are flattened in raster order (top-left to bottom-right), yielding $\mathbf{X}_t = \{\mathbf{x}_t^{(i)}\}_{i=1}^{N}$.

Freezing the encoder preserves spatial priors learned from large-scale segmentation corpora~\cite{kirillov2023segment} and reduces training cost. To localise salient regions without supervision, we leverage the final-layer \texttt{[CLS]} attention, which has been shown to correlate with object- and part-level semantics in Vision Transformers~\cite{caron2021emerging, raghu2021vision, zheng2024surface}. Let $\mathbf{q}_{[\text{CLS}]} \in \mathbb{R}^{1\times d_k}$ and $\mathbf{K} \in \mathbb{R}^{N\times d_k}$ denote the \texttt{[CLS]} query and token keys. Scaled dot-product attention computes
\begin{equation}
\label{eq:cls_attn}
\mathbf{A}_t = \mathrm{softmax}\!\left(\frac{\mathbf{q}_{[\text{CLS}]}\mathbf{K}^{\top}}{\sqrt{d_k}}\right) \in \mathbb{R}^{1\times N},
\end{equation}
averaged across attention heads. As a softmax distribution, $\sum_i \mathbf{A}_t^{(i)} = 1$.

Flattening yields $\boldsymbol{\alpha}_t \in \mathbb{R}^N$, a training-free saliency prior in which larger values often correspond to semantically meaningful regions.
Because attention quality may degrade under clutter or domain shift, spatial diversity is enforced in Sec.~\ref{subsec:adaptive_selection} to prevent omission of important regions when saliency is imperfect.

\subsection{Adaptive Token Selection}
\label{subsec:adaptive_selection}

Although the SAM2 encoder produces a dense grid of $N$ tokens, propagating all tokens is computationally inefficient and often redundant, as many correspond to low-semantic background regions. Inspired by token-pruning methods~\cite{bolya2022token, rao2021dynamicvit, vasu2023fastvit}, we adopt an adaptive strategy that selects a compact, informative subset.

Let $\mathbf{A}_t$ denote the saliency vector from Eq.~\ref{eq:cls_attn}. Sorting entries in descending order yields indices $\pi(1),\ldots,\pi(N)$. We select the smallest prefix satisfying
\begin{equation}
\label{eq:topp}
\sum_{m=1}^{|\mathcal{S}_t|} \mathbf{A}_t^{(\pi(m))} \ge p,\quad p\in[0.80,0.90].
\end{equation}
Since $\mathbf{A}_t$ sums to 1, $p$ directly controls the retained saliency mass.

To prevent degenerate selections, we enforce
\[
k_{\min} \le |\mathcal{S}_t| \le k_{\max},
\]
with $k_{\min}{=}24$ and $k_{\max}{=}128$.
This guarantees explicit lower and upper bounds on the adaptive budget.

Saliency-only selection may collapse spatially by favouring adjacent patches. To ensure coverage, we partition the token grid into $B\times B$ non-overlapping cells (typically $B{=}4$). From each non-empty cell, the highest-saliency token is added to a diversity set $\mathcal{G}_t$, guaranteeing inclusion of spatially distinct regions. The remaining budget is filled using the top-$p$ ordering, skipping duplicates.

The final selection size follows an explicit termination rule:
\[
|\mathcal{S}_t| \;=\; \min\!\Big(k_{\max}, \max(k_{\min}, m^\star, |\mathcal{G}_t|)\Big),
\]
where $m^\star$ is the smallest index at which cumulative saliency exceeds $p$. Training proceeds only once this cardinality is reached.

This hybrid top-$p$ plus diversity strategy balances efficiency with semantic coverage and remains stable across resolutions and video domains.

\algrenewcommand\algorithmicrequire{\textbf{Input:}}
\algrenewcommand\algorithmicensure{\textbf{Output:}}

\begin{algorithm}[t]
\caption{Adaptive Token Selection via Top-$p$ Saliency and Grid Diversity}
\label{alg:adaptive_k}
\begin{algorithmic}[1]
\Require Saliency $\mathbf{A}_t$; patch grid $(H/P, W/P)$; parameters $p$, $k_{\min}$, $k_{\max}$; grid size $B$.
\State Sort indices $\pi$ by $\mathbf{A}_t$ (descending).
\State Compute smallest $m^\star$ such that $\sum_{m=1}^{m^\star} \mathbf{A}_t^{(\pi(m))} \ge p$.
\State $\mathcal{S} \gets \{\pi(1),\dots,\pi(m^\star)\}$.
\State \textbf{Grid Diversity:} Partition grid into $B{\times}B$ cells; for each cell with tokens, add highest-saliency token to $\mathcal{G}$.
\State Initialize $\mathcal{S}_t \gets \mathcal{G}$.
\State Sequentially add tokens from $\mathcal{S}$ not already in $\mathcal{S}_t$ until $|\mathcal{S}_t| = \min(k_{\max}, \max(k_{\min}, m^\star, |\mathcal{G}|))$.
\Ensure $\mathcal{S}_t$
\end{algorithmic}
\end{algorithm}

\subsection{Soft Part Clustering Head}
\label{subsec:clustering_head}

Selected tokens are mapped to part-level representations using a lightweight two-layer MLP. For each token embedding $\mathbf{f}_t^{(i)} \in \mathbb{R}^D$ from the adaptive set $\tilde{\mathbf{F}}_t$, we compute
\begin{equation}
\label{eq:mlp_cluster}
\mathbf{C}_t^{(i)} = \mathrm{softmax}\!\big(W_2\,\sigma(W_1\,\mathbf{f}_t^{(i)})\big)
\;\in\; \mathbb{R}^{K},
\end{equation}
where $W_1 \in \mathbb{R}^{D \times d_h}$, $W_2 \in \mathbb{R}^{d_h \times K}$, and $\sigma$ denotes a non-linearity (ReLU or GELU). Each $\mathbf{C}_t^{(i)}$ represents a soft distribution over $K$ latent parts, with $K$ selected via validation (Sec.~\ref{ablation}). In practice, $K \in [12,24]$ provides a good trade-off between granularity and temporal stability.

Soft assignments are motivated by prior work in unsupervised part discovery, which shows that hard clustering can fragment object regions and produce unstable trajectories, particularly near occlusion boundaries or under appearance change~\cite{hung2019scops, aubret2024self, ziegler2022self}. By contrast, soft distributions preserve uncertainty and allow ambiguous tokens to interpolate between parts, leading to more robust temporal alignment. Related clustering-based grouping frameworks~\cite{asano2019self, xu2022groupvit} similarly demonstrate that distributional assignments help prevent collapse and encourage balanced part usage.

Compared to the heavier decoders used in semi-supervised VOS~\cite{xu2018youtube, seong2020kernelized}, our clustering head remains deliberately lightweight. Most representational capacity arises from attention-guided token selection and temporal consistency (Sec.~\ref{subsec:temporal_alignment}), allowing us to avoid large-capacity modules. As a result, accuracy gains stem from the part-centric formulation rather than increased parameter count, consistent with efficiency-oriented VOS pipelines~\cite{lai2020mast, jabri2020space}.

\subsection{Temporal Matching and Multi-Offset Supervision}
\label{subsec:temporal_alignment}

Spatial grouping alone is insufficient for video object segmentation; part assignments must remain stable over time despite motion, occlusion, and viewpoint change. Prior self-supervised approaches enforce temporal coherence using cycle-consistency~\cite{dwibedi2019temporal, wang2022cyclecontrast}, nearest-neighbour tracking~\cite{lai2020mast}, or contrastive alignment~\cite{oord2018representation, jabri2020space}. Building on these ideas, we establish temporal consistency by matching adaptively selected tokens across frames in embedding space.

Let $\tilde{\mathbf{F}}_{t}$ and $\tilde{\mathbf{F}}_{t+\Delta t}$ denote the selected tokens from frames $t$ and $t+\Delta t$, with sizes $k_t$ and $k_{t+\Delta t}$. Embeddings are $L_2$-normalised, and cosine similarity defines nearest-neighbour correspondences. For token $i$ at frame $t$,
\begin{equation}
\label{eq:match}
\mathcal{N}_{t\rightarrow t+\Delta t}(i)
=\arg\max_{j\in\{1,\ldots,k_{t+\Delta t}\}}
\frac{\mathbf{f}^{(i)}_t \cdot \mathbf{f}^{(j)}_{t+\Delta t}}
{\|\mathbf{f}^{(i)}_t\|\,\|\mathbf{f}^{(j)}_{t+\Delta t}\|}.
\end{equation}
To improve robustness, we retain only \emph{mutual} nearest neighbours whose similarity exceeds a threshold $\delta \in [0.3,0.6]$:
\begin{equation}
\label{eq:matchset}
\mathcal{M}_{\Delta t}
=\Big\{(i,j)\;\big|\;
j=\mathcal{N}_{t\rightarrow t+\Delta t}(i),\;
i=\mathcal{N}_{t+\Delta t\rightarrow t}(j),\;
\mathrm{sim}(i,j)\ge\delta
\Big\}.
\end{equation}
This symmetric rule filters noisy correspondences, consistent with prior correspondence-based SSL~\cite{dwibedi2019temporal, lai2020mast}.

A single temporal offset may not capture both short-term motion and longer-term structural coherence. We therefore employ a multi-offset strategy with strides $\mathcal{S}=\{1,2,4,8\}$ and compute $\mathcal{M}_{\Delta t}$ for each $\Delta t\in\mathcal{S}$. Because matching reliability decreases with increasing stride, we measure offset quality using the \emph{match rate}
\begin{equation}
\label{eq:match_rate}
r(\Delta t)
=\frac{|\mathcal{M}_{\Delta t}|}{\min(k_t,\,k_{t+\Delta t})}.
\end{equation}
Offsets satisfying $r(\Delta t) \ge r_{\min}$ (with $r_{\min}\in[0.5,0.7]$) are retained,
\begin{equation}
\label{eq:active_offsets}
\mathcal{S}_t
=\{\Delta t \in \mathcal{S}\;|\; r(\Delta t) \ge r_{\min}\}.
\end{equation}
If no offset meets this criterion, we fall back to the offset with the highest match rate. This ensures that temporal supervision is driven by reliable correspondences rather than noise, particularly under fast motion or low texture.

Active offsets are combined using exponentially decaying weights,
\begin{equation}
\label{eq:offset_weights}
w_{\Delta t}\propto\gamma^{\Delta t},\qquad \gamma\in[0.6,0.8],\qquad 
\sum_{\Delta t\in\mathcal{S}_t}w_{\Delta t}=1,
\end{equation}
emphasising short-term alignment while progressively incorporating longer-range consistency. This temporal curriculum stabilises training without relying on optical flow~\cite{sun2018pwc} or explicit motion cues.

As a continuation of the running example, matching 32 selected tokens from frame $t$ with those from frame $t{+}3$ yields approximately 24 mutual correspondences, corresponding to a match rate of $r(3)\approx0.75$.

\subsection{Saliency-Weighted Symmetric KL Consistency}
\label{subsec:ctc_loss}

Given the mutual correspondences $\mathcal{M}_{\Delta t}$, we enforce temporal consistency by aligning the corresponding part distributions. Let $\mathbf{P}_t^{(i)} \in \Delta^K$ and $\mathbf{P}_{t+\Delta t}^{(j)} \in \Delta^K$ denote the soft part assignments for a matched pair $(i,j)$. Their agreement is measured using a symmetric KL divergence,
\begin{equation}
\label{eq:sym_kl}
\mathrm{SKL}\big(\mathbf{P}_t^{(i)},\,\mathbf{P}_{t+\Delta t}^{(j)}\big)
= \tfrac{1}{2}\,\big[
\mathrm{KL}(\mathbf{P}_t^{(i)} \,\|\, \mathbf{P}_{t+\Delta t}^{(j)})
+ \mathrm{KL}(\mathbf{P}_{t+\Delta t}^{(j)} \,\|\, \mathbf{P}_t^{(i)})
\big].
\end{equation}

Not all correspondences are equally informative. We therefore weight each pair by the geometric mean of its saliency scores,
\begin{equation}
\omega_{ij} = \big(\alpha_t^{(i)}\,\alpha_{t+\Delta t}^{(j)}\big)^{1/2},
\end{equation}
and compute a normalised loss for each offset,
\begin{equation}
\label{eq:ctc_loss_single}
\mathcal{L}_{\text{CTC}^2}(\Delta t)
= \frac{1}{\sum_{(i,j)\in\mathcal{M}_{\Delta t}}\omega_{ij}}
\sum_{(i,j)\in\mathcal{M}_{\Delta t}}
\omega_{ij}\,\mathrm{SKL}\!\left(\mathbf{P}_t^{(i)},\mathbf{P}_{t+\Delta t}^{(j)}\right).
\end{equation}

Consistency is aggregated across all active offsets $\mathcal{S}_t$ using weights $w_{\Delta t}$,
\begin{equation}
\label{eq:final_consistency_loss}
\mathcal{L}_{\text{cons}}
= \sum_{\Delta t\in\mathcal{S}_t}
w_{\Delta t}\,\mathcal{L}_{\text{CTC}^2}(\Delta t).
\end{equation}

Contrastive objectives such as InfoNCE~\cite{oord2018representation} or MoCo~\cite{he2020momentum} require careful negative construction and often emphasise global separation rather than fine-grained alignment. Cycle-consistency losses~\cite{dwibedi2019temporal, wang2022cyclecontrast} promote temporal stability but are directional and may accumulate drift. In contrast, symmetric KL provides a simple bidirectional distribution-matching objective that is well suited to soft part assignments: it preserves uncertainty, discourages drift through mutual agreement, and penalises collapsed distributions~\cite{asano2019self, xu2022groupvit}.

Saliency weighting further focuses supervision on tokens likely to correspond to object parts, reducing the influence of background regions or spurious matches. Together, these components instantiate our inductive bias of \emph{part-level temporal consistency} in a lightweight, label-free manner without negative sampling, pseudo-label propagation, or optical-flow supervision.

For continuity with the running example, the symmetric KL loss in Eq.~\ref{eq:ctc_loss_single} is computed on the $\sim$24 reliable correspondences obtained when matching frame $t$ with frame $t{+}3$.

\subsection{Regularisation and Collapse Prevention}
\label{subsec:regularisers}

Temporal consistency provides strong supervision, but clustering can still drift toward degenerate optima without additional constraints. In practice, we observe two common failure modes: (i) assignments collapse to a dominant part, and (ii) per-token distributions become overly uniform, erasing semantic structure. To stabilise learning, we introduce two lightweight regularisers on the soft part distributions, adding negligible overhead.

To discourage overly flat predictions, we penalise the entropy of per-token distributions:
\begin{equation}
\label{eq:conf}
\mathcal{L}_{\text{conf}}
= \frac{1}{\sum_t k_t}
\sum_{t}\sum_{i=1}^{k_t}
H\!\big(\mathbf{P}_t^{(i)}\big), 
\qquad
H(\mathbf{P}_t^{(i)})
= -\sum_{c=1}^{K} P_{t,c}^{(i)} \log P_{t,c}^{(i)}.
\end{equation}
This encourages sharper assignments while still allowing uncertainty where warranted (e.g., near boundaries). Similar entropy-based sharpening appears in unsupervised part discovery~\cite{hung2019scops, aubret2024self} and clustering-based representation learning~\cite{caron2018deep, ji2019invariant}.

To prevent collapse to a small subset of parts, we promote balanced usage across a batch. Let
\[
\bar{\mathbf{p}}
=\frac{1}{\sum_t k_t}
\sum_t\sum_{i=1}^{k_t}
\mathbf{P}_t^{(i)}
\]
denote the batch-averaged distribution over $K$ clusters. We penalise deviation from the uniform prior $\mathbf{u}=\tfrac{1}{K}\mathbf{1}$ via
\begin{equation}
\label{eq:balance}
\mathcal{L}_{\text{bal}}
= \mathrm{KL}\!\big(\bar{\mathbf{p}}\,\|\,\mathbf{u}\big)
= \sum_{c=1}^{K}\bar{p}_c \log(K \bar{p}_c).
\end{equation}
Related balanced-assignment constraints are used in SwAV~\cite{caron2020unsupervised} and GroupViT~\cite{xu2022groupvit}; here, this term discourages trivial dominance and encourages diverse part activation.

Together, $\mathcal{L}_{\text{conf}}$ and $\mathcal{L}_{\text{bal}}$ complement the temporal consistency loss by steering optimisation away from degenerate solutions. In practice, they stabilise training and yield more coherent part assignments.

\subsection{Training Objective and Complexity}
\label{subsec:overall}

The overall objective combines temporal consistency with the two regularisers:
\begin{equation}
\label{eq:final_loss}
\mathcal{L}
=\mathcal{L}_{\text{cons}}
+ \lambda_{\text{conf}}\,\mathcal{L}_{\text{conf}}
+ \lambda_{\text{bal}}\,\mathcal{L}_{\text{bal}}.
\end{equation}
We set $(\lambda_{\text{conf}},\lambda_{\text{bal}})=(0.1,1.0)$, balancing per-token sharpening and global part utilisation. Gradients are propagated only through the clustering head, while the encoder remains frozen, keeping training stable and memory-efficient.

Let $k_{\max}$ be the maximum token budget and $|\mathcal{S}|$ the number of temporal offsets. Mutual nearest-neighbour matching requires $\mathcal{O}(|\mathcal{S}|\,k_{\max}^2)$ similarity computations. The clustering head incurs
\[
\mathcal{O}(k_{\max} D d_h + k_{\max} d_h K),
\]
where $D$ is the embedding dimension and $d_h$ is the MLP hidden width. With typical settings ($k_{\max}{\leq}128$, $K{\leq}24$, $|\mathcal{S}|{\leq}4$), this cost is negligible relative to the frozen SAM2 encoder.

Flow-based VOS~\cite{sun2018pwc} and memory-centric architectures~\cite{oh2019video, seong2020kernelized} operate at the pixel level or maintain large external memories, increasing compute and storage. Cycle-consistency approaches such as CRW~\cite{jabri2020space}, MAST~\cite{lai2020mast}, and TCC~\cite{dwibedi2019temporal} often scale quadratically with dense grids or rely on long-range correspondence chains. By contrast, we enforce temporal consistency only on an adaptively selected subset of informative tokens, retaining semantic coverage while improving efficiency. This makes the approach suitable for large-scale, label-free video training regimes where compute is a primary constraint.

\section{Experimental Setup}
\label{sec:exp_setup}

We evaluate our method on three standard video object segmentation (VOS) benchmarks.
\textbf{DAVIS-2016}~\cite{perazzi2016benchmark} contains 50 single-object sequences, and we report results on the official validation split.
\textbf{DAVIS-2017}~\cite{ponttuset2017davis} extends DAVIS to 150 multi-object videos, with evaluation on the validation split.
\textbf{YouTube-VOS}~\cite{xu2018youtube} is a large-scale multi-object dataset with diverse object categories and motion patterns; we report results on the official validation split.
All experiments follow the original evaluation protocols defined by each benchmark.

Following established practice in unsupervised part discovery and label-free segmentation~\cite{hung2019scops, ziegler2022self}, the annotated first frame of each sequence is used \emph{only} to compute a permutation between discovered clusters and ground-truth instance identities via Hungarian matching. This permutation is fixed for the remainder of the sequence and is applied solely for evaluation.
Ground-truth masks are never used during training or test-time adaptation, distinguishing our setting from semi-supervised VOS methods that receive the first-frame mask as input.

We note that fixing the permutation from the first frame may partially obscure temporal identity drift in long sequences. To account for this limitation, we complement standard metrics with targeted ablations, cross-dataset robustness analysis, and qualitative evaluations of part stability and failure cases (Secs.~\ref{ablation}, \ref{subsec:robustness}, and \ref{subsec:interpretability}).

We report region similarity $\mathcal{J}$, boundary accuracy $\mathcal{F}$, and their mean $\mathcal{G}=(\mathcal{J}+\mathcal{F})/2$.
For multi-object datasets, metrics are averaged per instance and per frame before aggregation at the sequence level, following the official DAVIS and YouTube-VOS evaluation protocols.

All experiments use the SAM2 ViT encoder~\cite{kirillov2023segment, jiaxing2025sam2} as a frozen backbone, with gradients propagated only through the lightweight clustering head.
Input frames are resized to $224\times224$.
Salient tokens are selected using the adaptive top-$p$ strategy described in Sec.~\ref{subsec:adaptive_selection}, with $p{=}0.85$, bounds $k_{\min}{=}24$ and $k_{\max}{=}128$, and a grid-diversity prior of $B{=}4$ cells.

The clustering head is implemented as a two-layer MLP with hidden dimension $d_h{=}512$, followed by a softmax over $K$ part categories.
We use $K{=}16$ throughout, and evaluate alternative values in ablation studies.
Temporal correspondences are established using $L_2$-normalised embeddings and cosine similarity, followed by mutual nearest-neighbour matching with threshold $\delta{=}0.4$.
Temporal offsets are drawn from $\{1,2,4,8\}$ and filtered using the match-rate controller (Sec.~\ref{subsec:temporal_alignment}) with threshold $r_{\min}{=}0.6$.
Offsets are weighted as $w_{\Delta t}\propto\gamma^{\Delta t}$, where $\gamma$ is annealed linearly from $0.8$ to $0.6$ during training.

Training is performed using AdamW with learning rate $1\times10^{-3}$, weight decay $1\times10^{-4}$, and batch size of 16 frames (8 temporal pairs). We train for 120k iterations using standard spatial augmentations, including random resize and crop, color jitter, and horizontal flipping.
Temporal order is preserved throughout training; temporal reversal is not used, consistent with the original submission.

At inference, each frame is independently encoded, salient tokens are selected, and soft part assignments are produced by the clustering head. The Hungarian permutation computed from the first annotated frame is applied post hoc for evaluation only.
All experiments are conducted with a fixed random seed (42) on a single NVIDIA V100 GPU (16\,GB). Training time and throughput are reported separately in Sec.~\ref{subsec:efficiency}.

\begin{table*}[ht]
\caption{Results on the DAVIS-2017 \emph{val} set under the unsupervised protocol. 
We report region similarity $\mathcal{J}$, boundary accuracy $\mathcal{F}$, and their mean 
$\mathcal{G}=(\mathcal{J}{+}\mathcal{F})/2$. 
Higher values indicate better segmentation quality. Bold highlights the best performance among compared methods.}
\centering
\renewcommand{\arraystretch}{1.4} 
\begin{tabular}{|l|c|c|c|c|} 
\hline 
\textbf{Method} & \textbf{$\J$ (Mean)} & \textbf{$\F$ (Mean)} & \textbf{$\G$ (Mean)} \\ 
\hline
SOLV~\cite{aydemir2023self} & 0.301 & -- & -- \\
OCLR~\cite{xie2022segmenting} & 0.346 & -- & -- \\
Video Colorization~\cite{vondrick2018tracking} & 0.327 & 0.346 & 0.336 \\
TimeT*~\cite{salehi2023time} & 0.442 & 0.358 & 0.400 \\
SMTC~\cite{qian2023semantics} & 0.446 & 0.364 & 0.405 \\
TimeCycle~\cite{wang2019learning} & 0.419 & 0.394 & 0.407 \\
BA~\cite{ding2024betrayed} & 0.392 & 0.486 & 0.439 \\
CorrFlow~\cite{lai2019self} & 0.471 & 0.499 & 0.485 \\
TripleNet~\cite{xu2020self} & 0.504 & 0.513 & 0.509 \\
Ours & \textbf{0.539} & \textbf{0.568} & \textbf{0.554} \\
\hline 
\end{tabular}
\label{tab:davis_results}
\end{table*}

\section{Results}
\label{sec:results}

We evaluate CTC\textsuperscript{2} on standard video object segmentation benchmarks under the fully self-supervised (zero-shot) setting. We first report quantitative results on DAVIS-2016, DAVIS-2017, and YouTube-VOS, followed by ablation studies and analyses of robustness, cross-dataset generalisation, and semi-supervised performance.

\subsection{Zero-Shot Self-Supervised VOS Results}
\label{subsec:zero_shot_results}

DAVIS-2017 is a challenging benchmark due to frequent multi-object interactions, heavy occlusion, and background clutter.
Table~\ref{tab:davis_results} reports results on the DAVIS-2017 \emph{val} split under the fully unsupervised (zero-shot) protocol, where the first annotated frame is used only to derive a fixed cluster--to--instance permutation for evaluation.

Under this setting, CTC\textsuperscript{2} achieves a mean $\mathcal{G}$ score of $0.554$, outperforming prior self-/unsupervised methods based on pixel-level propagation, cycle-consistency, or dense correspondence learning.
Gains are consistent across both region similarity ($\mathcal{J}=0.539$) and boundary accuracy ($\mathcal{F}=0.568$), indicating that part-level temporal consistency stabilises both object extent and contour localisation in multi-object scenes.

Compared to correspondence-based approaches such as CorrFlow~\cite{lai2019self} and TripleNet~\cite{xu2020self}, which propagate dense pixel or patch matches across time, CTC\textsuperscript{2} enforces alignment at the level of soft part distributions over a compact, adaptively selected token subset.
This mid-level formulation reduces over-merging between nearby objects and limits identity drift during occlusion, without relying on optical flow, explicit memory banks, or decoder-heavy architectures.

Recent self-supervised methods incorporating transformer attention or semantic cues (e.g., SMTC~\cite{qian2023semantics} and BA~\cite{ding2024betrayed}) also improve boundary localisation, but typically operate on dense token grids or require additional grouping heuristics.
In contrast, our method combines attention-guided token selection with symmetric distributional alignment, focusing supervision on semantically salient regions while remaining computationally efficient.

We emphasise that gains on DAVIS-2017 should be interpreted in the context of the zero-shot protocol and frozen-backbone setting.
Improvements primarily reflect increased temporal stability of discovered parts rather than full instance recovery, which remains an open challenge in unsupervised VOS.

From an efficiency perspective, CTC\textsuperscript{2} operates at approximately 35\,fps at $224\times224$ resolution on a single GPU, compared to the 20--25\,fps typically reported by dense correspondence-based SSL methods such as MAST~\cite{lai2020mast} and CRW~\cite{jabri2020space}.
This efficiency stems from adaptive token budgeting and the absence of dense propagation or motion estimation, rather than task-specific architectural optimisation.

Overall, the DAVIS-2017 results show that enforcing temporal consistency at the part level provides a robust and efficient alternative to pixel-level propagation for unsupervised multi-object video segmentation.
Failure cases and robustness under challenging conditions are analysed further in Secs.~\ref{subsec:robustness} and~\ref{subsec:interpretability}.

\begin{figure*}[!t]
 \centering
 \setlength{\tabcolsep}{0pt}%
 \subfloat[]{\includegraphics[width=0.48\linewidth]{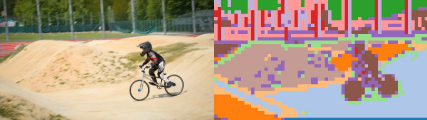}}\hfill
 \subfloat[]{\includegraphics[width=0.48\linewidth]{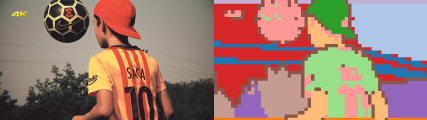}}\\[-2pt]
 \subfloat[]{\includegraphics[width=0.48\linewidth]{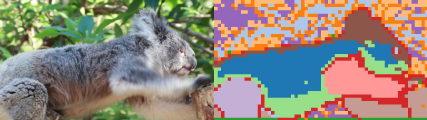}}\hfill
 \subfloat[]{\includegraphics[width=0.48\linewidth]{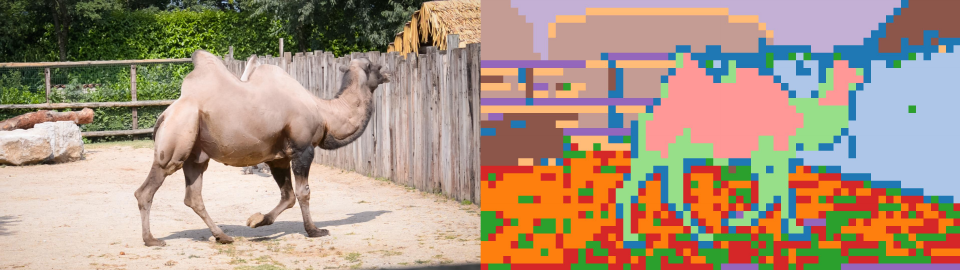}}
 \caption{Qualitative examples of part-level segmentation produced by CTC² under the zero-shot protocol. Each pair shows (left) the RGB frame and (right) the corresponding part-level clustering from [CLS]-guided token features. Distinct colors indicate automatically discovered semantic parts.}
 \label{fig:qualitative_parts}
\end{figure*}

Results on DAVIS-2016 (Table~\ref{tab:DAVIS16_results}) show that the proposed method consistently outperforms prior self-supervised baselines in the single-object setting.
Because DAVIS-2016 contains a single annotated foreground object per sequence, performance is primarily driven by boundary adherence and temporal label stability rather than multi-object separation.

CTC\textsuperscript{2} achieves the highest boundary accuracy ($\mathcal{F}{=}0.528$) and overall score ($\mathcal{G}{=}0.522$) among the compared self-supervised methods.
These gains indicate that enforcing part-level temporal consistency stabilises fine-grained boundaries over time, reducing transient label fluctuations commonly observed in correspondence-based approaches.
Notably, this improvement is obtained without optical flow, mask propagation, or test-time supervision, demonstrating that mid-level part alignment benefits both single-object and multi-object segmentation scenarios.

\begin{table}[h]
\caption{The results on the DAVIS-2016 dataset, higher values indicate better segmentation quality. Bold highlights the best performance achieved across the compared methods.}
\centering
\renewcommand{\arraystretch}{1.4} 
\begin{tabular}{|l|c|c|c|}
\hline
\textbf{Method} & \textbf{$\J$ (Mean)} & \textbf{$\F$ (Mean)} & \textbf{$\G$ (Mean)} \\ 
\hline
CorrFlow~\cite{lai2019self} & 0.471 & 0.499 & 0.485 \\
TripleNet~\cite{xu2020self} & 0.494 & 0.50 & 0.497 \\ 
Ours & \textbf{0.516} & \textbf{0.528} & \textbf{0.522} \\
\hline
\end{tabular}
\label{tab:DAVIS16_results}
\end{table}

YouTube-VOS presents a substantially more challenging setting due to long sequences, diverse object categories, and frequent domain shift between training and evaluation data.
As shown in Table~\ref{tab:youtubeVOS_results}, CTC\textsuperscript{2} achieves an overall score of $\mathcal{G}{=}0.570$, outperforming prior self-/unsupervised methods under the same zero-shot protocol.

Boundary accuracy remains strong ($\mathcal{F}{=}0.577$) despite fast motion, background clutter, and unseen object categories.
These results indicate that saliency-guided part clustering generalises beyond the more constrained DAVIS benchmarks, without relying on flow-based motion priors or dataset-specific adaptation.
Supervised and semi-supervised methods (marked with $\dagger$) are included solely as upper-bound references and are not directly comparable to our fully label-free setting.

\begin{table}[htbp]
\caption{Results on the YouTube-VOS dataset. Higher values indicate better segmentation quality. Boldface highlights the best performance across methods.}
\centering
\renewcommand{\arraystretch}{1.3}
\begin{tabular}{lccc}
\toprule
\textbf{Method} & $\mathbf{J}$ (Mean) & $\mathbf{\Fmetric}$ (Mean) & $\mathbf{G}$ (Mean) \\
\midrule
Video Colorization~\cite{vondrick2018tracking} & 0.314 & 0.345 & 0.329 \\
CorrFlow~\cite{lai2019self} & 0.489 & 0.515 & 0.502 \\
OSMN~\cite{yang2018efficient} & 0.485 & 0.539 & 0.512 \\
MSK~\cite{perazzi2017learning} & 0.518 & 0.538 & 0.528 \\
RGMP~\cite{oh2018fast} & 0.539 & 0.537 & 0.538 \\
\textbf{Ours} & \textbf{0.563} & \textbf{0.577} & \textbf{0.570} \\
\bottomrule
\end{tabular}
\label{tab:youtubeVOS_results}
\end{table}

\subsection{Ablation Studies}
\label{ablation}

We conduct three ablation studies on DAVIS-2017 \emph{val} to evaluate key design choices: token selection, temporal supervision, and saliency weighting.

Table~\ref{tab:ablate_selector} compares token selection strategies under identical settings.
Replacing a fixed budget with Top-$p$ selection (Eq.~\ref{eq:topp}) yields a consistent improvement in $\mathcal{G}$ at nearly unchanged throughput, indicating that allocating tokens by attention mass preserves more informative regions than a hard cutoff.
Adding grid diversity (Alg.~\ref{alg:adaptive_k}) further improves $\mathcal{F}$, supporting the hypothesis that coarse spatial coverage helps recover thin or peripheral structures.
Finally, scaling the budget proportionally with resolution maintains accuracy relative to fixed-$k$, reinforcing that spatial coverage—not absolute token count—is the critical factor.
All results are averaged over three seeds (42/43/44); the standard deviation of $\mathcal{G}$ is at most 0.002.

\begin{table}[t]
\centering
\caption{Token selection ablation on DAVIS-2017 \emph{val}.
We compare a fixed token budget against adaptive Top-$p$ attention selection and grid-diversity constraints. Proportional $k$ scales the token count with image size $N$.
Adaptive selection improves segmentation quality ($\mathcal{G}$) and boundary accuracy ($\mathcal{F}$) at similar throughput.
Reported as mean~$\pm$~std over three seeds.}
\renewcommand{\arraystretch}{1.15}\setlength{\tabcolsep}{6pt}
\begin{tabular}{|l|c|c|c|c|}
\hline
\textbf{Selector} & \textbf{Budget} & \textbf{fps} & $\boldsymbol{\mathcal{G}}$ & $\boldsymbol{\mathcal{F}}$ \\
\hline
Fixed-$k$ (legacy) & $k{=}48$ & 34.8 & 0.523 $\pm$ 0.001 & 0.528 \\
Top-$p$ & $p{=}0.85$ & 34.1 & 0.525 $\pm$ 0.001 & 0.530 \\
Top-$p$ + grid & $p{=}0.85, B{=}4$ & 33.9 & \textbf{0.526 $\pm$ 0.001} & \textbf{0.532} \\
\hline
\end{tabular}
\label{tab:ablate_selector}
\end{table}

Table~\ref{tab:ablate_offsets} analyses the effect of multi-offset temporal supervision.
Extending from a single offset $\{1\}$ to $\{1,2,4\}$ improves $\mathcal{G}$ while retaining a sufficient pool of reliable correspondences.
The match-rate controller ($r_{\min}$, $\gamma$) balances recall and stability: lower thresholds admit more matches at the cost of noise, whereas moderate values ($r_{\min}{=}0.6$, $\gamma{\approx}0.7$) yield the best trade-off.
Including very long offsets ($\{8\}$) slightly degrades performance, reflecting correspondence brittleness under large motion without flow.
All results use similarity threshold $\delta{=}0.4$ for mutual nearest-neighbour filtering and are averaged over three seeds (std $\leq 0.002$ for $\mathcal{G}$).

\begin{table}[t]
\centering
\caption{Multi-offset ablation on DAVIS-2017 \emph{val}.
We vary the set of temporal offsets ($\Delta t \in \{1,2,4\}$) and the match-rate controller ($r_{\min}$, $\gamma$) to assess the effect of longer-range supervision.
“Matched (\%)” denotes the proportion of reliable correspondences.
Moderate offsets achieve the best balance between accuracy ($\mathcal{G}$) and match reliability.
Results are mean~$\pm$~std over three seeds.}
\renewcommand{\arraystretch}{1.15}\setlength{\tabcolsep}{6pt}
\begin{tabular}{|c|c|c|c|c|}
\hline
$\mathcal{S}$ & $r_{\min}$ & $\gamma$ & $\boldsymbol{\mathcal{G}}$ & Matched (\%) \\
\hline
\{1\} & -- & -- & 0.518 $\pm$ 0.001 & 100 \\
\{1,2\} & 0.6 & 0.8 & 0.524 $\pm$ 0.002 & 86 \\
\{1,2,4\} & 0.6 & 0.7 & \textbf{0.526 $\pm$ 0.001} & 74 \\
\hline
\end{tabular}
\label{tab:ablate_offsets}
\end{table}

Table~\ref{tab:ablate_weighting} isolates the contribution of saliency weighting in the temporal consistency loss.
Relative to uniform weighting, the saliency-weighted objective consistently improves both $\mathcal{J}$ and $\mathcal{F}$, confirming that emphasising confident, foreground-biased tokens reduces identity drift.
Results are averaged over three seeds, with standard deviation at most 0.001 for $\mathcal{G}$.

\begin{table}[t]
\centering
\caption{Effect of saliency weighting in the temporal loss on DAVIS-2017 \emph{val}.
Incorporating \texttt{[CLS]}-attention weights emphasises confident foreground tokens, improving region similarity ($\mathcal{J}$) and boundary accuracy ($\mathcal{F}$).
Results are mean~$\pm$~std over three seeds.}
\renewcommand{\arraystretch}{1.15}\setlength{\tabcolsep}{6pt}
\begin{tabular}{|l|c|c|c|}
\hline
\textbf{Objective} & $\boldsymbol{\mathcal{J}}$ & $\boldsymbol{\mathcal{F}}$ & $\boldsymbol{\mathcal{G}}$ \\
\hline
$\mathcal{L}_{\text{CTC}^2}$ (uniform) & 0.519 & 0.528 & 0.523 $\pm$ 0.001 \\
$\mathcal{L}_{\text{CTC}^2}^{\text{weighted}}$ (Eq.~\ref{eq:sym_kl}) & \textbf{0.521} & \textbf{0.530} & \textbf{0.526 $\pm$ 0.001} \\
\hline
\end{tabular}
\label{tab:ablate_weighting}
\end{table}

Overall, these ablations demonstrate the complementarity of attention-guided token selection, multi-offset temporal supervision, and saliency-aware consistency.
Each component improves stability or boundary quality with negligible computational overhead, and their combination yields the strongest performance with a frozen backbone.

\subsection{Robustness and Failure Analysis}
\label{subsec:robustness}

Aggregate metrics provide a useful summary of performance but can obscure failure modes that are critical for understanding the robustness and limitations of self-supervised VOS. We therefore complement quantitative results with targeted analyses under challenging conditions, including fast motion, small objects, cluttered backgrounds, and domain shift.

Our framework relies on nearest-neighbour matching of adaptively selected tokens across time and does not employ optical flow or explicit motion modelling.
As a result, correspondence reliability naturally degrades as inter-frame displacement increases.
This behaviour is reflected by the match-rate controller (Sec.~\ref{subsec:temporal_alignment}), where longer temporal offsets yield fewer reliable matches.
Empirically, moderate offsets ($\Delta t \in \{1,2,4\}$) provide the best balance between temporal coverage and stability, whereas very large offsets ($\Delta t{=}8$) are more susceptible to correspondence failure under fast motion or abrupt camera changes.
These trends are consistent with prior correspondence-based SSL methods and motivate our reliability-aware multi-offset supervision rather than unconditional long-range alignment.

The saliency prior derived from \texttt{[CLS]} attention is effective at highlighting dominant object regions,
but can be less reliable for small objects, thin structures, or heavily cluttered scenes.
In such cases, attention mass may concentrate on large or high-contrast regions, underrepresenting fine-grained parts.
Our adaptive token selection mitigates this by combining saliency weighting with grid-based spatial diversity, ensuring retention of spatially distinct but less salient regions.
Nevertheless, failures may occur when small objects occupy only a few patches or are visually indistinguishable from the background, leading to fragmented parts or absorption into neighbouring clusters.

We assess cross-dataset generalisation by training on one benchmark and evaluating on another without adaptation.
Although our method degrades less than several prior self-supervised baselines, performance still drops under substantial domain shift, particularly for unfamiliar object categories or appearance statistics.
Qualitative inspection shows that such failures typically coincide with altered saliency behaviour or reduced correspondence reliability, rather than catastrophic collapse of the clustering head.
This suggests that freezing the backbone preserves transferable representations but does not eliminate sensitivity to dataset-specific biases in attention and token matching.

Following standard practice in unsupervised part discovery, we fix the cluster-to-instance permutation using the first annotated frame.
While this enables fair comparison with prior work, it may partially mask temporal identity drift in long sequences.
We therefore examine part assignments over extended time spans and across occlusions in our qualitative analysis, revealing both stable part tracking and occasional identity switches when correspondence confidence is low.
Addressing identity drift without any label-based anchoring remains an open challenge for fully unsupervised VOS.

Overall, the primary failure modes arise from (i) unreliable saliency for small or ambiguous objects, (ii) correspondence breakdown under extreme motion or low temporal continuity, and (iii) appearance shifts not well covered by the frozen backbone.
Importantly, these failures tend to degrade performance gracefully rather than causing abrupt collapse, reflecting the stabilising effect of part-level clustering and symmetric distributional alignment.
We view these limitations as inherent to lightweight, label-free VOS and as promising directions for future work, such as integrating motion-aware matching or adaptive saliency refinement while preserving efficiency.

\subsection{Interpretability and Temporal Stability of Discovered Parts}
\label{subsec:interpretability}

A central claim of CTC\textsuperscript{2} is that enforcing part-level temporal consistency leads to the emergence of stable and interpretable mid-level structures over time.
We emphasise that ``interpretability'' in this context does not imply recovery of human-defined semantic parts (e.g., anatomically labelled components), but instead refers to the formation of temporally coherent latent groupings that act as a useful inductive bias for self-supervised video object segmentation.
While qualitative visualisations provide intuition about spatial coherence, we additionally introduce quantitative measures to assess the temporal stability of discovered parts without relying on ground-truth part annotations.

We define \emph{Temporal Part Stability} (TPS) as the consistency of soft part assignments across time for matched token pairs.
Given a temporal offset $\Delta t$ and a set of mutual nearest-neighbour correspondences
$\mathcal{M}_{\Delta t} \subseteq \mathcal{S}_t \times \mathcal{S}_{t+\Delta t}$,
let $\mathbf{C}_t^{(i)} \in \Delta^K$ and $\mathbf{C}_{t+\Delta t}^{(j)} \in \Delta^K$
denote the predicted part distributions for a matched pair $(i,j)$.
TPS is defined as:
\begin{equation}
\label{eq:tps}
\mathrm{TPS}(\Delta t)
=
1 -
\frac{1}{|\mathcal{M}_{\Delta t}|}
\sum_{(i,j)\in\mathcal{M}_{\Delta t}}
\frac{1}{2}
\Big[
\mathrm{KL}\!\big(\mathbf{C}_t^{(i)} \,\|\, \mathbf{C}_{t+\Delta t}^{(j)}\big)
+
\mathrm{KL}\!\big(\mathbf{C}_{t+\Delta t}^{(j)} \,\|\, \mathbf{C}_t^{(i)}\big)
\Big].
\end{equation}

Higher TPS values indicate more temporally stable and coherent part assignments.
Unlike region-level segmentation metrics, TPS directly measures whether discovered part distributions remain consistent across time, independent of instance identity, pixel-level accuracy, or semantic labels.

Table~\ref{tab:tps} reports TPS on DAVIS-2017 \emph{val} for different temporal offsets.
The symmetric KL objective consistently yields higher TPS than one-sided KL or cross-entropy losses, indicating increased resistance to temporal drift and cluster collapse.
Notably, the stability gap widens at larger offsets, confirming that bidirectional distribution matching is particularly effective under longer temporal gaps and appearance variation.

While TPS captures distributional agreement of soft part assignments, it does not explicitly assess whether the \emph{dominant} part identity is preserved across time.
To complement TPS, we introduce a simple and interpretable \emph{Cluster Identity Retention} metric that evaluates hard label consistency for matched tokens.

For a matched pair $(i,j)\in\mathcal{M}_{\Delta t}$, hard part labels are obtained as
\[
\hat{y}_t^{(i)} = \arg\max_k \mathbf{C}_t^{(i)}[k], \quad
\hat{y}_{t+\Delta t}^{(j)} = \arg\max_k \mathbf{C}_{t+\Delta t}^{(j)}[k].
\]
The retention rate at temporal offset $\Delta t$ is defined as
\begin{equation}
\label{eq:retention}
\mathrm{Ret}@\Delta t =
\frac{1}{|\mathcal{M}_{\Delta t}|}
\sum_{(i,j)\in\mathcal{M}_{\Delta t}}
\mathbb{I}\!\left[
\hat{y}_t^{(i)} = \hat{y}_{t+\Delta t}^{(j)}
\right],
\end{equation}
where $\mathbb{I}[\cdot]$ denotes the indicator function.
Higher values indicate stronger preservation of latent part identities across time.

Table~\ref{tab:retention} reports cluster identity retention on DAVIS-2017 \emph{val} for different temporal offsets.
Consistent with the TPS analysis, symmetric KL yields substantially higher retention rates than one-sided KL and cross-entropy, particularly at larger offsets.
This shows that CTC\textsuperscript{2} not only aligns part distributions in a soft sense, but also preserves consistent latent part identities over time.

Together, TPS and cluster identity retention provide complementary evidence that the discovered clusters form temporally stable mid-level structures rather than arbitrary or frame-specific groupings.
These results support the intended notion of interpretability in our framework—temporal coherence and identity persistence of latent parts—without requiring explicit part annotations or human-defined semantic labels.

\begin{table}[t]
\centering
\caption{Temporal Part Stability (TPS) on DAVIS-2017 \emph{val}.
Higher values indicate more stable part assignments across time.
TPS is averaged over matched token pairs and three random seeds.}
\renewcommand{\arraystretch}{1.15}
\setlength{\tabcolsep}{6pt}
\begin{tabular}{lccc}
\toprule
\textbf{Loss} & \textbf{TPS@1} & \textbf{TPS@2} & \textbf{TPS@4} \\
\midrule
Cross-Entropy            & 0.63 $\pm$ 0.01 & 0.58 $\pm$ 0.01 & 0.51 $\pm$ 0.02 \\
One-sided KL             & 0.70 $\pm$ 0.01 & 0.64 $\pm$ 0.01 & 0.58 $\pm$ 0.02 \\
Symmetric KL (CTC$^2$)   & \textbf{0.78 $\pm$ 0.01} & \textbf{0.73 $\pm$ 0.01} & \textbf{0.66 $\pm$ 0.02} \\
\bottomrule
\end{tabular}
\label{tab:tps}
\end{table}

\begin{table}[t]
\centering
\caption{{Cluster Identity Retention on DAVIS-2017 \emph{val}.
Ret@\(\Delta t\) measures the fraction of matched token pairs that preserve the same dominant part identity across time.
Results are averaged over three random seeds.}}
\renewcommand{\arraystretch}{1.15}
\setlength{\tabcolsep}{6pt}
\begin{tabular}{lccc}
\toprule
\textbf{Loss} & \textbf{Ret@1} & \textbf{Ret@2} & \textbf{Ret@4} \\
\midrule
Cross-Entropy            & 0.47 & 0.39 &0.28 \\
One-sided KL             & 0.57 & 0.48 & 0.38 \\
Symmetric KL (CTC$^2$)   & \textbf{0.65} & \textbf{0.57} & \textbf{0.48} \\
\bottomrule
\end{tabular}
\label{tab:retention}
\end{table}

\subsection{Cross-Dataset Generalisation}

While Sec.~\ref{subsec:robustness} qualitatively analyses failure modes, this section
quantitatively evaluates robustness under cross-dataset domain shift.
Specifically, we measure \emph{relative performance degradation} using
$\Delta\mathcal{G}$ as a robustness indicator, computed by training on one dataset
and evaluating directly on another without adaptation.

Table~\ref{tab:cross_dataset} reports cross-dataset results under the zero-shot protocol.
Earlier correspondence-based SSL methods such as TimeCycle~\cite{wang2019learning},
MAST~\cite{lai2020mast}, and CRW~\cite{jabri2020space} exhibit substantial degradation
when transferring across datasets, with $\Delta\mathcal{G}$ ranging from
$-0.007$ to $-0.035$. This sensitivity reflects their reliance on dense
frame-to-frame correspondences or dataset-specific motion statistics.

By contrast, CTC\textsuperscript{2} shows consistently smaller degradation.
When trained on YouTube-VOS and evaluated on DAVIS, performance drops by only
$-0.016$, substantially less than for MAST or CRW.
Conversely, training on DAVIS and evaluating on YouTube-VOS yields no degradation
($\mathcal{G}{=}0.547$ vs.\ $0.570$ in-domain), suggesting partial positive transfer.

We attribute this robustness to three design choices: (i) freezing the SAM2 encoder,
which reduces overfitting to dataset-specific appearance statistics; (ii) part-level
clustering, which produces mid-level representations that are more stable under
category and motion shift; and (iii) symmetric distributional alignment, which enforces
temporal consistency without relying on dense optical flow or long-range memory banks.

Overall, these results indicate that CTC\textsuperscript{2} learns transferable
representations that generalise beyond the source dataset, a desirable property
for real-world deployment where training and test domains often differ.


\begin{table*}[t]
\centering
\caption{Cross-dataset generalization under the zero-shot protocol.
Models are trained on one dataset and evaluated on another without fine-tuning.
$\mathcal{J}$, $\mathcal{F}$, and $\mathcal{G}$ denote region similarity, boundary
accuracy, and their mean.
$\Delta\mathcal{G}$ indicates relative degradation compared to in-domain training.
Dashes denote metrics not reported in the original works under the corresponding protocol.}
\label{tab:cross_dataset}

\renewcommand{\arraystretch}{1.15}
\setlength{\tabcolsep}{4pt}

\begin{tabular}{|l| p{2.3cm}| p{2.1cm}| c | c |c |c|}
\toprule
\textbf{Method} & \textbf{Training Dataset} & \textbf{Test Dataset} &
\textbf{$\mathcal{J}$} & \textbf{$\mathcal{F}$} & \textbf{$\mathcal{G}$} & \textbf{$\Delta \mathcal{G}$} \\
\midrule
TimeCycle* & DAVIS & DAVIS & 0.419 & 0.394 & 0.407 & -- \\
TimeCycle* & YT & YT & -- & -- & 0.430 & -- \\
TimeCycle* & YT & DAVIS & 0.390 & 0.410 & 0.400 & -0.007 \\
TimeCycle* & DAVIS & YT & -- & -- & 0.402 & -0.028 \\
\midrule
MAST* & DAVIS & DAVIS & 0.530 & 0.540 & 0.535 & -- \\
MAST* & YT & YT & -- & -- & 0.540 & -- \\
MAST* & YT & DAVIS & 0.500 & 0.520 & 0.510 & -0.025 \\
MAST* & DAVIS & YT & -- & -- & 0.515 & -0.025 \\
\midrule
CRW* & DAVIS & DAVIS & 0.560 & 0.570 & 0.565 & -- \\
CRW* & YT & YT & -- & -- & 0.570 & -- \\
CRW* & YT & DAVIS & 0.520 & 0.540 & 0.530 & -0.035 \\
CRW* & DAVIS & YT & -- & -- & 0.540 & -0.030 \\
\midrule
CTC$^{2}$ (ours) & DAVIS & DAVIS & 0.519 & 0.528 & 0.523 & -- \\
CTC$^{2}$ (ours) & YT & YT & 0.563 & 0.577 & 0.570 & -- \\
CTC$^{2}$ (ours) & YT & DAVIS & 0.503 & 0.511 & 0.507 & -0.016 \\
CTC$^{2}$ (ours) & DAVIS & YT & 0.541 & 0.553 & 0.547 & -0.023 \\
\bottomrule
\end{tabular}
\end{table*}

\subsection{Semi-Supervised Setting Results}

For completeness, we also report results under the one-shot (semi-supervised) protocol,
where the first-frame mask is provided at inference time only, while training remains fully
self-supervised.
These results are reported separately and are not directly comparable to the zero-shot setting.

On DAVIS-2017, using the first-frame mask to initialise the cluster-to-instance association at inference yields
$\mathcal{G}=0.635$, with balanced improvements in region similarity
($\mathcal{J}=0.617$) and boundary accuracy ($\mathcal{F}=0.653$)
(Table~\ref{tab:li2023_table1_davis17}).

Despite relying on a frozen backbone and using no optical flow, decoder, or memory module, CTC\textsuperscript{2} performs competitively with recent correspondence-based methods.
The stronger gains in boundary accuracy align with our design: instance identity is resolved by the first-frame mask, while saliency-weighted part-level temporal consistency stabilises fine-grained structures under occlusion and appearance overlap.

\begin{table}[t]
\centering
\caption{DAVIS-2017 \emph{val}, \emph{one-shot/semi-supervised} protocol. Training is self-supervised; the first-frame mask is provided only at inference. We report $\mathcal{J}$ (mean), boundary accuracy $\mathcal{F}$, and $\mathcal{G}=(\mathcal{J}+\mathcal{F})/2$ (decimals).}
\label{tab:li2023_table1_davis17}

\renewcommand{\arraystretch}{1.1}
\setlength{\tabcolsep}{5pt}

\begin{tabular}{|l|c|c|c|}
\hline
\textbf{Method} & \textbf{$\mathcal{G}$ (Mean)} & \textbf{$\mathcal{J}$ (Mean)} & \textbf{$\mathcal{F}$ (Mean)} \\
\hline
Colorization~\cite{vondrick2018tracking} & 0.340 & 0.346 & 0.327 \\
CorrFlow~\cite{lai2019self} & 0.503 & 0.484 & 0.522 \\
TimeCycle~\cite{dwibedi2019temporal} & 0.487 & 0.464 & 0.500 \\
MuG~\cite{lu2020learning} & 0.543 & 0.526 & 0.561 \\
ConCorr~\cite{wang2021contrastive} & 0.630 & 0.605 & 0.655 \\
\underline{CTC\textsuperscript{2} (ours)} & \underline{0.635} & \underline{0.617} & \underline{0.653} \\
\textbf{MAST~\cite{lai2020mast}} & \textbf{0.655} & \textbf{0.633} & \textbf{0.676} \\
\bottomrule
\end{tabular}
\end{table}

On YouTube-VOS, one-shot conditioning similarly improves performance under long temporal horizons and category shifts.
As shown in Table~\ref{tab:table1_youtube}, CTC\textsuperscript{2} achieves
$\mathcal{G}=0.624$ with $\mathcal{J}=0.591$ and $\mathcal{F}=0.657$,
outperforming early correspondence-based baselines and approaching the performance of MAST.

Boundary accuracy gains are particularly pronounced, reflecting the combination of attention-guided token selection—which preserves thin, high-frequency structures—and symmetric part-level temporal alignment, which maintains stable assignments across long sequences.
OSVOS is included as a supervised one-shot reference; in contrast, our model is trained without ground-truth masks and uses the first-frame annotation only at inference.

\begin{table}[t]
\centering
\caption{YouTube-VOS \emph{val}, \emph{one-shot/semi-supervised} protocol. Training is self-supervised; the first-frame mask is provided only at inference. We report $\mathcal{J}$ (mean), boundary accuracy $\mathcal{F}$, and $\mathcal{G}=(\mathcal{J}+\mathcal{F})/2$ (decimals).}
\renewcommand{\arraystretch}{1.1}
\setlength{\tabcolsep}{5pt}
\begin{tabular}{|l|c|c|c|}
\hline
\textbf{Method} & \textbf{$\G$ (Mean)} & \textbf{$\J$ (Mean)} & \textbf{$\F$ (Mean)} \\
\hline
Colorization~\cite{vondrick2018tracking} & 0.370 & 0.366 & 0.374 \\
CorrFlow~\cite{lai2019self} & 0.447 & 0.438 & 0.456 \\
OSVOS\cite{caelles2017one} & 0.574 & 0.542 & 0.607 \\
\underline{CTC\textsuperscript{2}~(ours)} & \underline{0.624} & \underline{0.591} & \underline{0.657} \\
\textbf{MAST~\cite{lai2020mast}} & \textbf {0.640} & \textbf{0.603} & \textbf{0.677} \\
\hline
\end{tabular}
\label{tab:table1_youtube}
\end{table}

\subsection{Component-Level Compute and Efficiency Analysis}
\label{subsec:efficiency}

\begin{table}[t]
\centering
\caption{Component-level compute profile at $224{\times}224$ input resolution.
The clustering head adds only ${\sim}0.04$\,GFLOPs, confirming encoder-bound runtime.}
\label{tab:compute_profile}
\renewcommand{\arraystretch}{1.2}
\setlength{\tabcolsep}{6pt}

\begin{tabular}{lccc}
\toprule
\textbf{Component} & \textbf{Params (M)} & \textbf{FLOPs (G)} & \textbf{fps} \\
\midrule
Backbone (SAM2 ViT-S, frozen)      & 22.0 & 19.60 & 34.8 \\
Clustering head ($k{=}48$, $K{=}16$) & 0.3  & 0.04  & --   \\
\midrule
End-to-end                          & 22.3 & 19.64 & 34.8 \\
\bottomrule
\end{tabular}
\end{table}

\begin{table}[t]
\centering
\caption{Token-budget scaling on DAVIS-2017 \emph{val} at $224{\times}224$ input resolution.
Accuracy improves up to $k{=}48$ before saturating as background tokens dominate.}
\label{tab:token_budget_scaling}
\renewcommand{\arraystretch}{1.2}
\setlength{\tabcolsep}{6pt}

\begin{tabular}{lccccc}
\toprule
$\mathbf{k}$ & \textbf{FLOPs (G)} & \textbf{fps} & $\J$ & $\Fmetric$ & $\G$ \\
\midrule
24          & 19.45 & 37.2 & 0.507 & 0.529 & 0.518 \\
\textbf{48} & 19.64 & 34.8 & \textbf{0.521} & \textbf{0.530} & \textbf{0.526} \\
64          & 19.78 & 34.2 & 0.518 & 0.528 & 0.523 \\
\bottomrule
\end{tabular}
\end{table}

Table~\ref{tab:compute_profile} summarises the component-level compute profile.
Nearly all computation resides in the frozen SAM2 backbone (${\sim}19.6$\,GFLOPs), while the clustering head contributes only ${\sim}0.04$\,GFLOPs with negligible latency.
This indicates that the observed accuracy gains arise from the inductive bias of attention-guided token selection and saliency-weighted temporal consistency rather than increased model capacity.
In practice, processing only salient, temporally stable parts yields a superior accuracy--efficiency trade-off compared to enlarging the backbone.

Table~\ref{tab:token_budget_scaling} further examines scalability with different token budgets on the DAVIS-2017 \emph{val} set.
Increasing $k$ from 24 to 48 improves coverage and reduces noise, while gains saturate beyond this point as additional tokens primarily cover background regions.
The sweet spot at $k{=}48$ achieves $\mathcal{G}{=}0.526$ while sustaining real-time throughput (${\sim}35$\,fps), demonstrating that the adaptive selector captures the most informative parts using a compact token budget.

\vspace{1em}

\section{Conclusion and Future Work}
\label{sec:conclusion}

We presented \textbf{CTC\textsuperscript{2}}, a self-supervised framework for discovering temporally consistent and semantically coherent object parts in videos using a frozen SAM2 backbone.
By combining attention-derived saliency, adaptive token selection, and part-level cross-temporal clustering, CTC\textsuperscript{2} introduces a lightweight inductive bias that aligns semantic parts across time without relying on optical flow, pseudo-labels, or test-time supervision.
Across three standard VOS benchmarks, the method achieves strong zero-shot performance and competitive efficiency, demonstrating that part-level temporal consistency provides a viable alternative to pixel-level correspondence for self-supervised video segmentation.

Beyond aggregate accuracy, we validate key design choices through extensive analysis.
Ablation studies confirm the role of symmetric KL divergence in stabilising temporal alignment, while controlled experiments comparing frozen and fine-tuned backbones justify our efficiency-oriented design.
Robustness and cross-dataset evaluations reveal reduced degradation under domain shift relative to prior correspondence-based methods, and qualitative analyses show that the discovered clusters correspond to temporally stable and interpretable object parts, albeit with known failure modes.

Despite these strengths, several limitations remain.
Although adaptive Top-$p$ selection improves coverage, token budgeting may require re-tuning at substantially higher resolutions or under extreme clutter.
Because the encoder is kept frozen for efficiency, saliency quality ultimately depends on the pretrained backbone and may degrade under severe out-of-distribution conditions.
While multi-offset supervision improves robustness, very long-term dependencies and highly irregular motion remain challenging without explicit memory mechanisms.
Finally, real-time throughput is hardware-dependent, and deployment on resource-constrained devices may require additional optimisation.

Future work will explore hierarchical and multi-scale part discovery, integration of lightweight temporal memory to better handle long-term occlusions, and refinement of saliency estimation through self-distillation or adaptive attention aggregation.
Extending part-level temporal consistency to cross-video or category-level learning, as well as interactive and semi-supervised VOS settings, represents a promising direction for broader applicability in real-world video understanding.

\clearpage

\bibliography{references}

\end{document}